\documentclass[english]{article}

\usepackage[T1]{fontenc}
\usepackage[latin9]{inputenc}
\usepackage{geometry}
\usepackage{prettyref}
\usepackage{float}
\usepackage{graphicx}
\usepackage{setspace}
\usepackage[authoryear]{natbib}
\makeatletter

\@ifundefined{definecolor}
 {\usepackage{color}}{}

\definecolor{parametergray}{gray}{0.8}

\usepackage{babel}

\usepackage[%
colorlinks=true,linkcolor=blue,citecolor=blue,urlcolor=blue,filecolor=blue,%
pdffitwindow,%
]{hyperref}

\usepackage{colortbl}
\makeatletter\renewcommand{\@biblabel}[1]{\quad#1.}\makeatother
\usepackage{bm}

\date{}

\newrefformat{sec}{Section \ref{#1}}
\newrefformat{sub}{Section \ref{#1}}
\newrefformat{fig}{Figure \ref{#1}}
\newrefformat{eq}{Equation \ref{#1}}

\AtBeginDocument{
  
}

\makeatother

\usepackage{babel}
\begin{document}
\global\long\def\taudec{\tau_{\mathrm{dec}}}
\global\long\def\ms{\:\mathrm{ms}}
\global\long\def\Hz{\:\mathrm{Hz}}
\global\long\def\k{\:\mathrm{k}}
\global\long\def\m{\:\mathrm{m}}
\global\long\def\tsim{t_{\mathrm{sim}}}
\global\long\def\trun{t_{\mathrm{run}}}
\global\long\def\tbuild{t_{\mathrm{build}}}
\global\long\def\ttotal{t_{\mathrm{total}}}
\global\long\def\GB{\:\mathrm{GB}}

\title{Closed loop interactions between spiking neural network and robotic
simulators based on MUSIC and ROS}

\author{Philipp Weidel$^{1*}$, Mikael Djurfeldt$^{2,1}$, Renato Duarte$^{1,3,4}$,
Abigail Morrison$^{1,4,5,6}$}

\maketitle
\noindent$^{1}$\parbox[t]{11cm}{Institute for Advanced Simulation
(IAS-6),\\
Theoretical Neuroscience \& \\
Institute of Neuroscience and Medicine (INM-6),\\
Computational and Systems Neuroscience \& \\
JARA BRAIN Institute I\\
Jülich Research Center and JARA\\
Jülich, Germany}\\[2mm]

\noindent$^{2}$\parbox[t]{11cm}{PDC Center for High Performance
Computing\\
KTH\\
Stockholm, Sweden}\\[2mm]

\noindent$^{3}$\parbox[t]{11cm}{Faculty of Biology\\
Albert-Ludwig University of Freiburg\\
Freiburg im Breisgau, Germany}\\[2mm]

\noindent$^{4}$\parbox[t]{11cm}{Bernstein Center Freiburg\\
Albert-Ludwig University of Freiburg\\
Freiburg im Breisgau, Germany}\\[2mm]

\noindent$^{5}$\parbox[t]{11cm}{Simulation Laboratory Neuroscience
--\\
Bernstein Facility for Simulation and Database Technology\\
Institute for Advanced Simulation\\
Jülich Aachen Research Alliance\\
Jülich Research Center\\
Jülich, Germany}\\[2mm]

\noindent$^{6}$\parbox[t]{11cm}{Institute of Cognitive Neuroscience\\
Faculty of Psychology\\
Ruhr-University Bochum\\
Bochum, Germany}\\[2mm]

\vfill{}

\noindent$^{*}$\parbox[t]{11cm}{Forschungszentrum Jülich GmbH

Institute of Neuroscience and Medicine (INM-6)

52425 Jülich, Germany

e-mail: p.weidel@fz-juelich.de}

\thispagestyle{empty}
\begin{abstract}
In order to properly assess the function and computational properties
of simulated neural systems, it is necessary to account for the nature
of the stimuli that drive the system. However, providing stimuli that
are rich and yet both reproducible and amenable to experimental manipulations
is technically challenging, and even more so if a closed-loop scenario
is required. In this work, we present a novel approach to solve this
problem, connecting robotics and neural network simulators. We implement
a middleware solution that bridges the Robotic Operating System (ROS)
to the Multi-Simulator Coordinator (MUSIC). This enables any robotic
and neural simulators that implement the corresponding interfaces
to be efficiently coupled, allowing real-time performance for a wide
range of configurations. This work extends the toolset available for
researchers in both neurorobotics and computational neuroscience,
and creates the opportunity to perform closed-loop experiments of
arbitrary complexity to address questions in multiple areas, including
embodiment, agency, and reinforcement learning.

\end{abstract}

\section{Introduction}

Studying a functional, biologically plausible neural network that
performs a particular task is highly relevant for progress in both
neuroscience and robotics. The major focus on this topic in the field
of robotics consists of using of neural networks of varying degrees
of complexity for controlling robots. So far, the majority of research
has focused on non-spiking, artificial neural networks for this task
\citep{quinonez_autonomous_2015,antonelo_event_2007,dasgupta_information_2013},
but there is considerable interest in investigating the capacities
of spiking neural networks.

In computational neuroscience, the study of simulated neural networks
is paramount to gain a better understanding of the processes underlying
learning/adaptation to complex environments and global behavior \citep[e.g. ][]{chorley2008closing}.
However, neural network simulators do not typically include functionality
for representing environments and sensory input. As a consequence,
most tasks used to test the function of a simulated neural network
are hardcoded to represent a highly specific task. This has the disadvantage
that virtual experiments are complex and time consuming to develop
and adapt. More importantly, tasks defined in this way are rather
artificial \citep{jitsev2012learning,fremaux_reinforcement_2013,potjans2011imperfect,legenstein2014ensembles,friedrich2015goal}.
Whereas there is certainly value in investigating very simplified
tasks and sensory representations, it is also vital to be able to
check that proposed neural architectures are capable of handling richer,
noisier and more complex scenarios. 

However, design of environments, representation of sensory states
and conversion of motor commands into movements are primary features
of robotic simulators. Therefore, it would be of great value to both
research fields if the powerful tools developed for robotic simulation
and spiking neural network simulation could be made to work together.
This would allow researchers from both fields to perform simulated
closed loop experiments with flexible experiment design, rich sensory
input, complex neuronal processing, and motor output.  The major
challenge lies in the fact that robotic simulators communicate via
continuous data streams, while neural network simulators communicate
with spike events (so as their biological counterparts). Thus, a principled
approach for bi-directional conversion is required.

So far there have been few attempts to address this. \citet{moren2015real}
describe a technical setup for a specific use case, in which a robot
is connected to a neural network model of early saccade movement simulated
with the Neural Simulation Tool \citep[NEST:][]{Gewaltig:NEST}. As
the robot is not in the same location as the cluster where the neural
simulation is running, the connection is realized via the internet
using an SSH tunnel and the Multi-Simulation Coordinator \citep[MUSIC:][]{djurfeldt2010run},
a middleware facilitating communication between different neural simulators
in one simulation. Using a minimal model and optimal number of cores,
the NEST simulation ran a factor of two slower than real time and
achieved an output response of $110-140$$\ms$. Although this study
provides a proof of concept for interaction between robotics and neural
simulators, it does not represent a general solution, in part due
to the limited implementation detail provided. 

A more general solution was presented by \citet{hinkel_domain-specific_2015},
who describe an interface between the Robotic Operating System \citep[ROS:][]{quigley2009ros}
and NEST. ROS is the most popular middleware for the robotic community,
with interfaces to many robotic simulators and also robotic hardware.
It allows the user to create their own simulated robots in great detail
using the Unified Robot Description Format. Similarly, NEST is one
of the most commonly used neural simulators for spiking point neuron
models. However, the proposed interface is on the basis of Python:
the neural network must be simulated in brief periods in an external
loop, reading out and communicating the spikes to the robotic simulator
at each iteration. Although no performance data are provided, the
overheads inherent in repeatedly stopping and starting NEST imply
severe performance limitations for this approach.

In this paper we present an alternative approach based on ROS and
MUSIC that is both general and efficient, enabling real-time performance
for up to hundreds of thousands of neurons on our hardware. Our approach
provides roboticists the opportunity to extend their work on neural
control into the realm of spiking neural networks using any neural
simulator implementing a MUSIC interface, including NEST or NEURON
\citep{hines2001neuron}. Conversely, our toolchain frees the researcher
from the constrictions of the hardcoding approach described above,
by enabling neural simulators to be connected to any of the myriad
robotic simulators implementing a ROS interface (including Gazebo%
\footnote{http://gazebosim.org/%
}, Morse%
\footnote{https://www.openrobots.org/wiki/morse/%
} or Webots%
\footnote{https://www.cyberbotics.com%
}). To demonstrate the capabilities of the toolchain, we implement
a Braitenberg Vehicle \citep{braitenberg1986vehicles} in which the
agent is simulated in Gazebo and the neural controller in NEST.

\subsection{Description of the toolchain}

The main purpose of the present toolchain is to capture a broad set
of use cases by connecting robotic with neural simulators in a generic
way, obtained by interfacing the two middlewares (ROS and MUSIC).
The interface is licensed under GPLv3 and available on Github%
\footnote{https://github.com/weidel-p/ros\_music\_adapter%
}. In addition to our own repository, we plan to distribute the toolchain
as a plug-in for MUSIC%
\footnote{https://github.com/INCF/MUSIC%
}.  

In this work, we propose a set of possible solutions to overcome the
problem of converting between continuous and spiking signals. In particular,
we investigate the performance of three different kinds of encoders
(NEF, regular rate coding, Poisson rate coding) and a linear readout
decoder. As the correct encoding mechanism is debatable and dependent
on the scientific question being addressed, in addition to our own
mechanisms, we provide extensibility such that researchers can implement
their own custom encoders, decoders and adapters in Python or C++.

\begin{figure}[h]
\begin{centering}
\includegraphics[width=1\columnwidth]{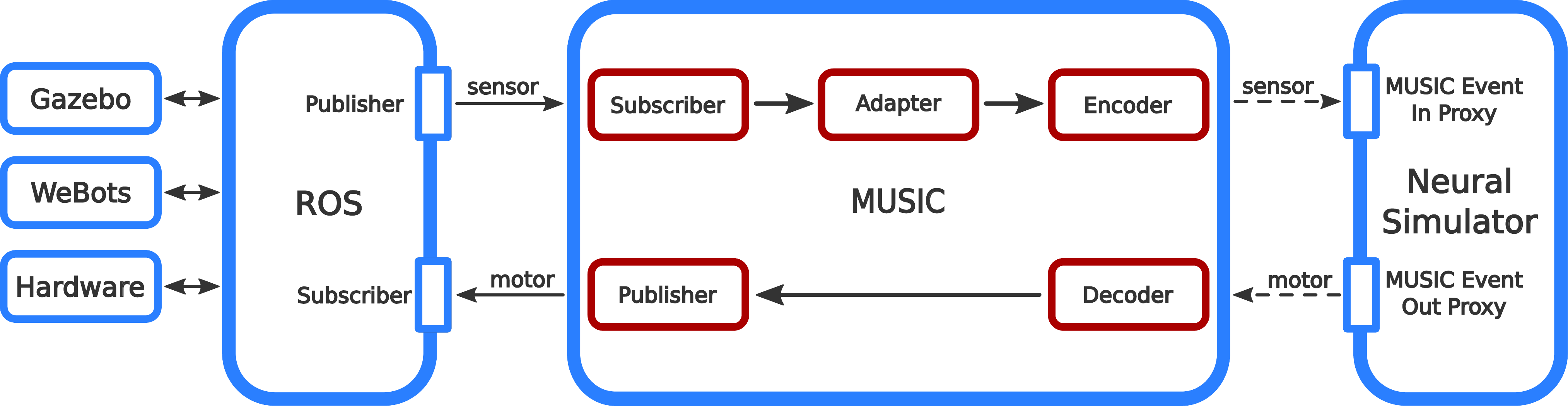}
\par\end{centering}

\caption{\label{fig:Block-diagram-of-the}Information flow in the ROS-MUSIC
toolchain. The boxes with red outlines depict our extensions to MUSIC
that facilitate the bidirectional conversion of continuous and spiking
signals (solid and dashed arrows, respectively)}
\end{figure}
To achieve these ends, our interface to ROS extends MUSIC by three
different kind of binaries: adapters, encoders and decoders (see \prettyref{fig:Block-diagram-of-the}).
Encoders and decoders are used to translate between spiking data
and continuous data while adapters can be used for pre-processing
data and connect to ROS. All binaries run in their own process and
solve only one specific purpose. This way, the interface is highly
modular and the implementation of custom adapters, encoders or decoders
is as simple as possible.

\section{Materials and Methods}

\subsection{The MUSIC configuration file}

The adapters, encoders, decoders and simulators introduced in the
toolchain are specified, connected and parameterized in the MUSIC
configuration file. Detailed information about the MUSIC configuration
file can be found in the MUSIC manual%
\footnote{http://software.incf.org/software/music%
} and examples of complete configuration files for this toolchain can
be found in the Github repository.

\subsection{Encoding, decoding and adaptation\label{sub:Encoding-and-decoding}}

The choice of encoder and decoder to convert continuous to spiking
signals is non-trivial, and potentially dependent on the scientific
question or task. Here, we describe three possible encoding mechanisms
and one decoding mechanism. We also describe a simple adapter to map
the dimenstionality of the input signal to the dimensionality of the
receiving network.

\subsubsection{Regular rate coding }

The simplest way to encode continuous data into spiking data is regular
rate coding. i. e. mapping the input signal to a time-dependent firing
rate variable. Each input dimension is encoded by a single neuron
by transforming the input signal, which is a continuous signal in
the range $[-1,1]$, into a time-dependent variable representing the
spike density of the encoding neuron. This is achieved by calculating
the interspike interval of each encoding neuron at time $t$ by:

\begin{equation}
ISI_{n}(t)=\frac{1}{v_{\mathrm{min}}+(v_{\mathrm{max}}-v_{\mathrm{min}})\frac{1+I_{n}\left(t\right)}{2}}\label{eq:rate}
\end{equation}
where $I_{n}\left(t\right)$ is the input which is mapped to neuron
$n$. The firing rate is scaled between $[v_{\mathrm{min}},v_{\mathrm{max}}]$
which are free parameters of the regular rate encoder. This encoding
mechanism leads to very regular spike patterns but has the clear advantage
of being computationally very efficient. Note that this approach does
not take into consideration biological neuron properties such as refractoriness,
but these could be included without much impact on the computational
efficacy.

\subsubsection{Poisson rate coding }

In this encoding scheme, we follow a similar implementation to that
described in the previous section, but introduce a stochastic component
in the generated spike trains, by transforming the input signal into
a time-dependent variable representing the intensity of a stochastic
point process with Poissonian statistics. This can be achieved by
using the inverse of the expression (\prettyref{eq:rate}) for interspike
interval in the previous section as the rate parameter for an exponential
ISI distribution: 

\begin{equation}
ISI_{n}(t)=-\frac{1}{v_{\mathrm{min}}+(v_{\mathrm{max}}-v_{\mathrm{min}})\frac{1+I_{n}\left(t\right)}{2}}\ln(r)
\end{equation}
where $r$ is a random number drawn from a uniform distribution in
$\left[0,1\right)$ and $I_{n}\left(t\right)$ is the input mapped
to neuron $n$. The firing rate is scaled between $[v_{\mathrm{min}},v_{\mathrm{max}}]$
which are free parameters of the Poisson rate-encoder. This encoding
mechanism produces spike trains with Poissonian statistics, similar
to those typically assumed to occur in central neurons in the mammalian
brain (see, for example, \citet{Calvin1967,Dayan2001} or \citet{Averbeck2009}
for a counter-argument), and therefore are thought to be a more realistic
approximation than regular spike trains. As \prettyref{sec:Results}
shows, the computational complexity of this approach is comparable
to that of regular rate encoding.

\subsubsection{Neural Engineering Framework}

The Neural Engineering Framework \citep[NEF:][]{eliasmith2004neural}
is a generic formalism to represent stimuli with neural ensembles
that allows a wide variety of functions to be realised by deriving
optimal projections between neural populations. In the simplest formulation,
NEF encodes a vector of continuous data $I$ as an input current to
an integrate-and-fire neuron whose activity is defined by

\begin{equation}
a_{n}(I)=G_{n}[\alpha_{n}\cdotp I+I_{n}^{\mathrm{bias}}]
\end{equation}
where $G$ represents a generic non-linear functional which is determined
by the neuron dynamics, $\alpha_{n}$ is the neuron's response preference,
or tuning curve, and $I_{n}^{\mathrm{bias}}$ a constant bias current
setting the base activation of neuron $n$. It is worth noting that
this encoding does not depend on the specific form of $G$ and any
neuron model can be used for this operation. The original signal $I$
can be adequately reconstructed (depending on the properties of G
and alpha) by linearly combining the activity $a$ of the encoded
representations

\begin{equation}
\hat{I}=\sum_{n}a_{n}(I)\phi_{n}
\end{equation}
where the decoding weights $\phi$ are obtained by linear regression
in order to minimize the reconstruction error.

\subsubsection{Linear decoder }

Having described possible options explored in this work to convert
continuous signals into discrete spike event trains, we now explore
the reverse operation. To transform the spiking activity of a given
neural ensemble to a continuous signal, which can be used, for example,
to provide motor commands to the simulated robot, we first perform
a low-pass filter of each spike train by convolving with a causal
(i.e. only defined for $t\geq0$) exponential kernel $k(t)=\exp\left(-t/\taudec\right)$
with time constant $\taudec$

\begin{equation}
a_{n}(t)=\sum_{i}k(t-t_{i,n})
\end{equation}
where $t_{i,n}$ represents the $i$-th spike time of neuron $n$.
The resulting activities at time $t$ can be linearly combined to
obtain a continuous output signal 

\begin{equation}
z_{k}=\sum_{n}a_{n}(t)\phi_{nk}
\end{equation}
where weights $\phi_{nk}$ define the contribution of ensemble neuron
$n$ to readout unit $k$.

\subsubsection{Signal adapter}

For the experiments carried out below we implement a simple adapter
that maps the $m$ dimensions of the input signal to the $n$ dimensions
of the receiving neurons. Note that this is only necessary for regular
and Poisson rate encoding; the NEF encoder already incorporates this
functionality by construction. For the performance measurements, each
receiving neuron receives all dimensions of the input signal. For
the Braitenberg vehicle, the input signal is split up into two hemispheres,
each mapped to one of the two controlling neurons. For more complex
experiments, a more sophisticated adapter can be implemented.

\subsection{Performance Measurements\label{sub:Performance-measurements}}

In order to be as close as possible to a real use-case, we simulate
the whole toolchain while measuring the performance capabilities of
the different parts. The agent used for the performance measurements
is a four wheeled, mobile robot simulated in Gazebo with an attached
virtual laser scanner. The laser scanner has $100$ beams with an
update rate of $20\Hz$ and a maximal range of $5\m$. We use the
\texttt{SkidSteerDrivePlugin} provided by Gazebo, which allows us
to steer the robot with a \texttt{ROS::Twist} message, updated with
a rate of $20\Hz$. During the measurements involving the NEF encoder,
we keep the parameters of the integrate-and-fire (IAF) neurons unchanged
and simulate these neurons with a resolution of $1\ms$ using exact
integration \citep{Rotter1999,Morrison2007,Hanuschkin2010}. For
decoding, we always use a linear readout where the spiking activity
is filtered with an exponential kernel with a resolution of $1\ms$.
The MUSIC adapters run with an update rate equal to the update rate
of the sensory input and motor command output. The measurements were
executed on one node of a transtec CALLEO $551$ cluster and were
averaged over five trials with a simulation-time of ten seconds.
  Measurements involving neural simulators were carried out with
NEST 2.8 \citep{eppler_2015_32969} and NEURON v6.2.

\subsubsection{Real-time factor\label{sub:RTF-measurements}}

The sensory input, the motor output and the processing of the sensory
data are all updated in parallel but asynchronously. For robotic applications
it is crucial that the execution time for processing input data in
each time step is less than or equal to the wall-clock time of each
time step, meaning the process is running in real-time. 

The real-time factor ($RTF$) is calculated by dividing the simulated
time, $\tsim$, by the wallclock time required to run the simulation,
$\trun$.

\begin{equation}
RTF=\frac{\tsim}{\trun}
\end{equation}
The total time the toolchain needs to perform a simulation can be
divided into different components
\begin{equation}
\ttotal=\tbuild+\trun+\epsilon
\end{equation}
where $\tbuild$ is the time the toolchain needs for initialization
(allocate memory etc), $\trun$ is the actual time the toolchain needs
to perform the simulation and $\epsilon$ is the time which is not
covered by $\tbuild$ and $\trun$ (i.e. garbage collection, freeing
memory, etc.). In our measurements we clearly separate the buildup
phase from the run-time phase by synchronizing the processes after
the initialization of the software using an MPI barrier. This allows
us to measure $\trun$ consistently over all processes. 

We use this measurement to investigate different aspects of the toolchain's
performance, as described below.

\paragraph{Dimensionality of the input }

The sensory input to a robot can be very complex and high dimensional.
The number of neurons needed to encode a multidimensional input strongly
depends on the encoding mechanism. In NEF, about $100$ IAF neurons
per dimension are used when encoding a stimulus in order to be able
to decode that stimulus with a root-mean-square error about $1\%$
\citep{eliasmith2004neural}. Using rate coding, the number of neurons
can be freely chosen. In the example of the Braitenberg Vehicle \citep{braitenberg1986vehicles},
we use only two neurons for encoding the complete sensory input (see
\prettyref{sub:Braitenberg}). 

To examine the scalability of the sensory input in our toolchain,
we measure how many encoding neurons for different encoders we can
simulate until the real-time performance breaks down. We use a binary
search to find these limits in an efficient way. When measuring the
performance with NEST and NEURON, each neuron instantiated by the
encoder is matched by a corresponding neuron in the neuronal simulator,
which receives the input spike train and repeats it to the decoder
as an identical spike train, thus implementing a minimal processing
network. In NEST this is realized by the \texttt{parrot\_neuron},
a neuron model that emits a spike for each spike received. In NEURON
the repeater neurons are of type \texttt{IntFire1} equipped with
very high input synaptic weights and zero refractory time, thus enforcing
the neuron to spike after each input spike. For the rate and Poisson
encoder, the encoder neurons have a low firing rate between $1$ and
$2\Hz$. Except for NEST ($7$ processes) and NEURON ($40$ processes),
each process in the toolchain runs on one dedicated process.

\paragraph{Bandwidth}

Apart from the computational limitations of simulating neurons, communication
is another potential bottleneck in this toolchain.  Parts of the
toolchain are communicating action potentials (events) between different
processes (i.e. from encoder to NEST). In order to investigate the
influence of the communication on the real-time performance, we use
a rate based encoder and measure the real-time factor as a function
of the firing rate of the encoding neurons. We choose the amount of
neurons close to the border of real-time capability, so that we can
see an immediate impact of the firing-rate on the performance of the
toolchain.

\paragraph{Latency\label{sub:Latency}}

The latency between sensory input and motor output can be crucial
for the robotic application. A fast reaction time is needed for many
applications (for example, catching a ball). In order to measure the
latency (or reaction time), we artificially disturb the sensory signal
of the robotic simulator, switching discontinuously from the minimum
to the maximum sensor range. A very simple encoder responds to this
change by beginning to produce spikes, which the neural simulator
receives over its MUSIC Event In Proxy and repeats its the MUSIC Event
Out Proxy (see \prettyref{fig:Block-diagram-of-the}). A decoder responds
to the arrival of this spike train by producing a motor command, which
is conveyed over ROS to the robotic simulator. The latency of the
tool chain is therefore the wall-clock time between the change in
the sensory signal and the reception of the motor signal.

\paragraph{Overhead}

For measuring the input scalability as described above, we use a MUSIC
time step equal to the ROS update rate. This minimizes the communication
overhead as only new data is communicated. As latency depends on the
MUSIC time step, we additionally investigate the communication overhead
of the toolchain. To do this, we determine the real-time factor of
the toolchain, using a simple rate encoding mechanism, whilst systematically
varying the MUSIC time step and the number of simulated neurons.

\section{Results\label{sec:Results}}

\begin{figure}
\begin{centering}
\includegraphics[width=1\columnwidth]{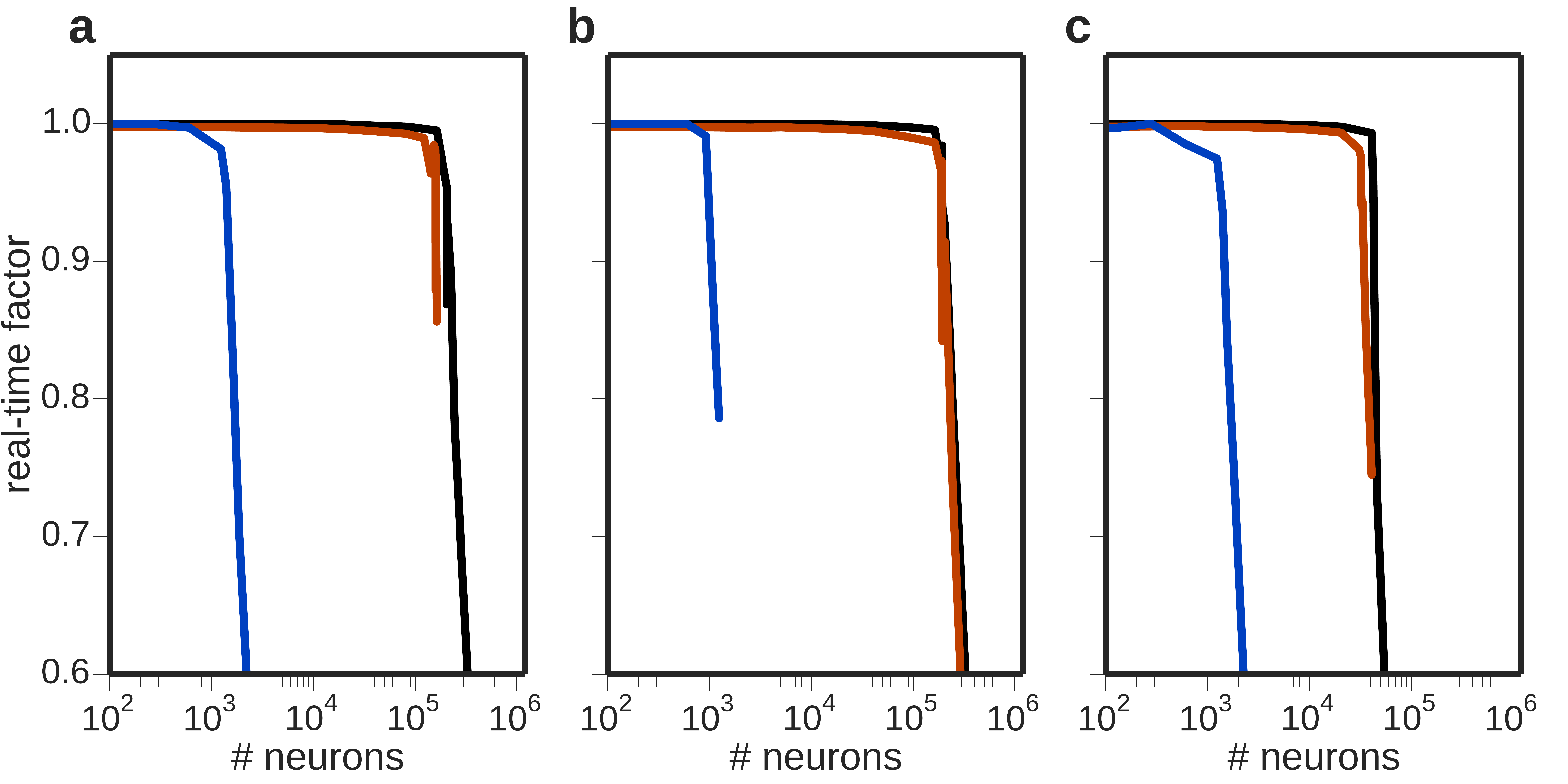}
\par\end{centering}

\caption{\label{fig:Scalability}Scalability of the toolchain. Real-time factor
as a function of the number of neurons used for encoding using different
encoders: \textbf{a)} rate encoding mechanism; \textbf{b)} Poissonian
encoding mechanism; \textbf{c)} NEF encoding mechanism. In each panel,
the black curve shows the real-time performance in the absence of
a neuronal simulator, the red curve shows the performance using NEST
and the blue curve the performance using NEURON. }
\end{figure}
\prettyref{fig:Scalability} shows the effect of the computational
load of the different encoding mechanisms on the real-time factor,
with and without neural simulators involved in the toolchain. For
the rate or Poissonian encoding mechanisms (\prettyref{fig:Scalability}\textbf{a,b}),
simulations of up to $150,000$ encoding neurons are possible in real-time
when using NEST or when the toolchain does not include a neural simulator.
When using NEURON, the real-time factor breaks down at about $1000$
neurons. For the NEF encoding mechanism (\prettyref{fig:Scalability}\textbf{c}),
simulations of up to $20,000$ encoding neurons are possible in real-time
when using NEST or no neural simulator in the toolchain. In this case
too, the performance when using NEURON breaks down at about $1000$
neurons.

These results demonstrate that the NEF encoder is computationally
more expensive than either the rate or Poissonian encoders, and that
NEURON is computationally expensive in this context, limiting the
real-time performance to $1000$ neurons regardless of which encoding
mechanism is used.

\begin{figure}
\begin{centering}
\includegraphics[width=1\columnwidth]{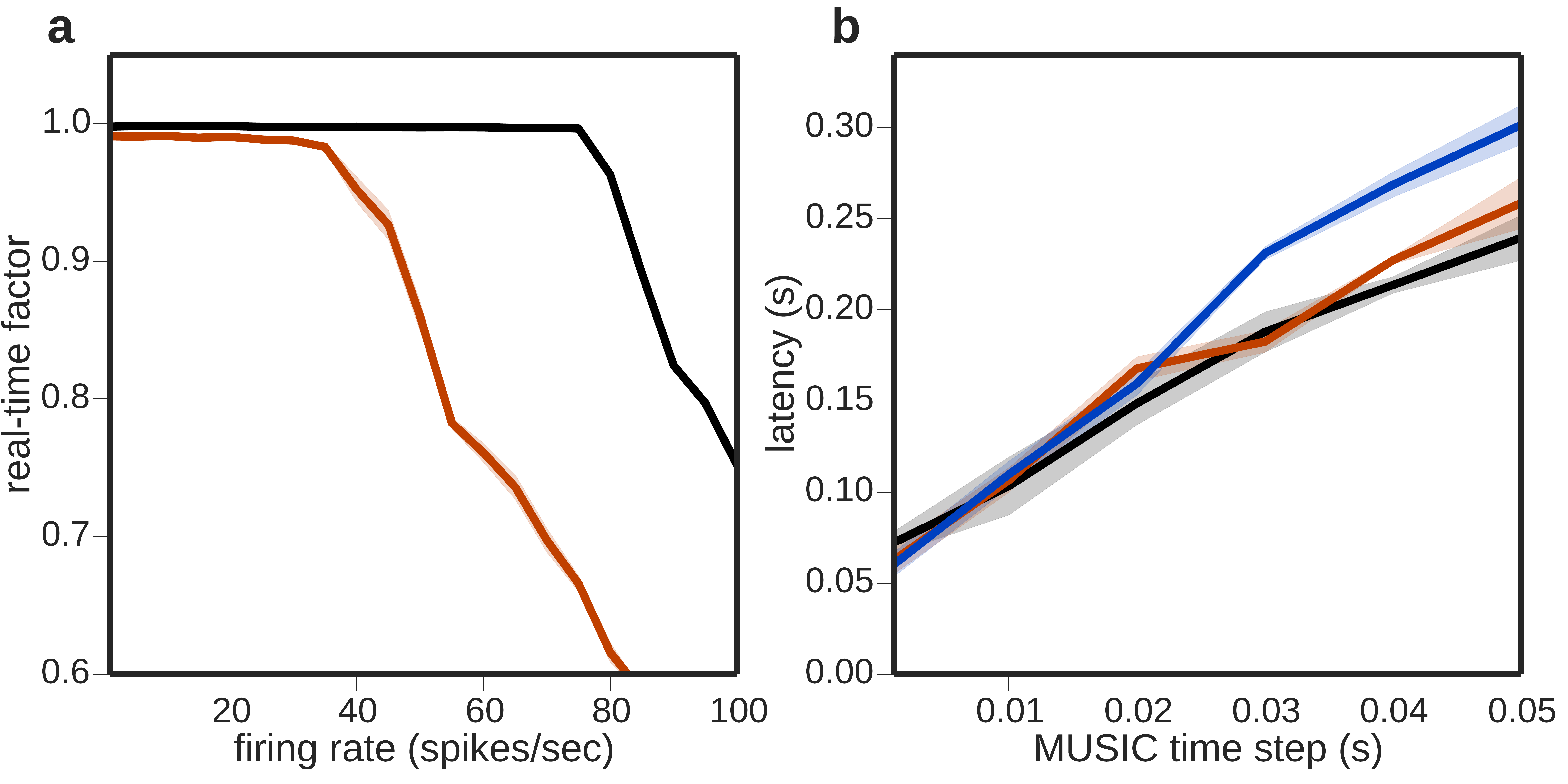}
\par\end{centering}

\caption{\label{fig:Bandwidth}Bandwidth limitations of the toolchain. \textbf{a)}
Real-time factor as a function of neuronal firing rate per neuron
for a regular rate encoder of $50,000$ neurons without a neural simulator
(black curve) and with NEST (red curve).  \textbf{b)} Latency as
a function of MUSIC time step without a neural simulator (black curve),
with NEST (red curve) and with NEURON (blue curve). }
\end{figure}
The breakdown of the real-time performance of the regular rate and
Poissonian  encoders happens at about the same amount of neurons,
which leads to the question what the actual bottleneck of the toolchain
is in these cases. Computational complexity is only one limiting factor
in this toolchain; another candidate is the communication.

But as \prettyref{fig:Bandwidth}\textbf{a} shows, the real-time factor
breaks down at about $40\Hz$ using $50,000$ encoding neurons. In
other words, the toolchain is able to communicate about $2,000,000$
spikes per second, which is more than was required for the simulations
measured in \prettyref{fig:Scalability}. Thus we can conclude firstly
that communication was not the bottleneck for the previous experiment,
and secondly that not just the number of the encoding neurons but
also their firing rate can be a limiting factor for the real-time
performance.

\prettyref{fig:Bandwidth}\textbf{b} shows the latency of the toolchain,
i.e. the difference in time between a change in sensory input and
receiving a motor command evoked by that change (see \prettyref{sub:RTF-measurements}).
The latency increases linearly with the MUSIC time step; a MUSIC time
step of $1\ms$, results in a latency of about $70\ms$, growing to
around $350\ms$ for a time step of $50\ms$ when using a neural simulator.
In general, the presence of a neural simulator in the toolchain increases
the latency slightly, independently of which simulator is chosen. 

\begin{figure}
\begin{centering}
\includegraphics[width=1\columnwidth]{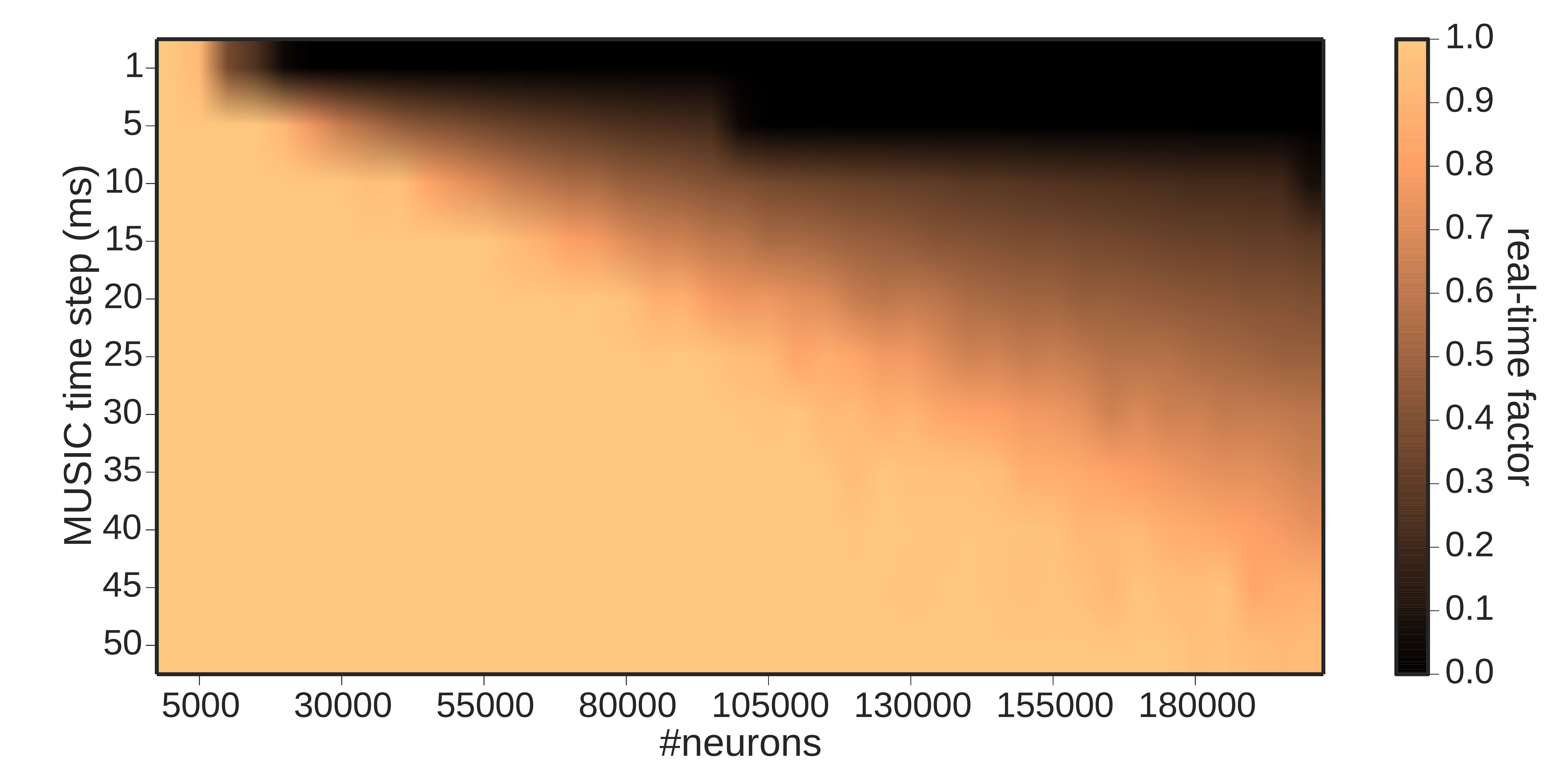}
\par\end{centering}

\caption{\label{fig:Overhead}Real-time factor as a function of the MUSIC time
step and number of neurons. }
\end{figure}
\prettyref{fig:Overhead} depicts the dependency of real-time capability
of the toolchain on the MUSIC time step. The border of real-time simulation
capability increases linearly with the MUSIC time step, from $10,000$
neurons for a time step of $1\ms$ to $185,000$ neurons for a $50\ms$
time step. However, as demonstrated in \prettyref{fig:Bandwidth}b,
the latency also increases with the MUSIC time step. From these results,
we conclude that the latency and the dimensionality of the input are
two conflicting properties, which have to be balanced for the specific
use case.

\subsection{Implementation of a Braitenberg Vehicle\label{sub:Braitenberg} }

\begin{figure}[H]
\begin{centering}
\includegraphics[width=1\columnwidth]{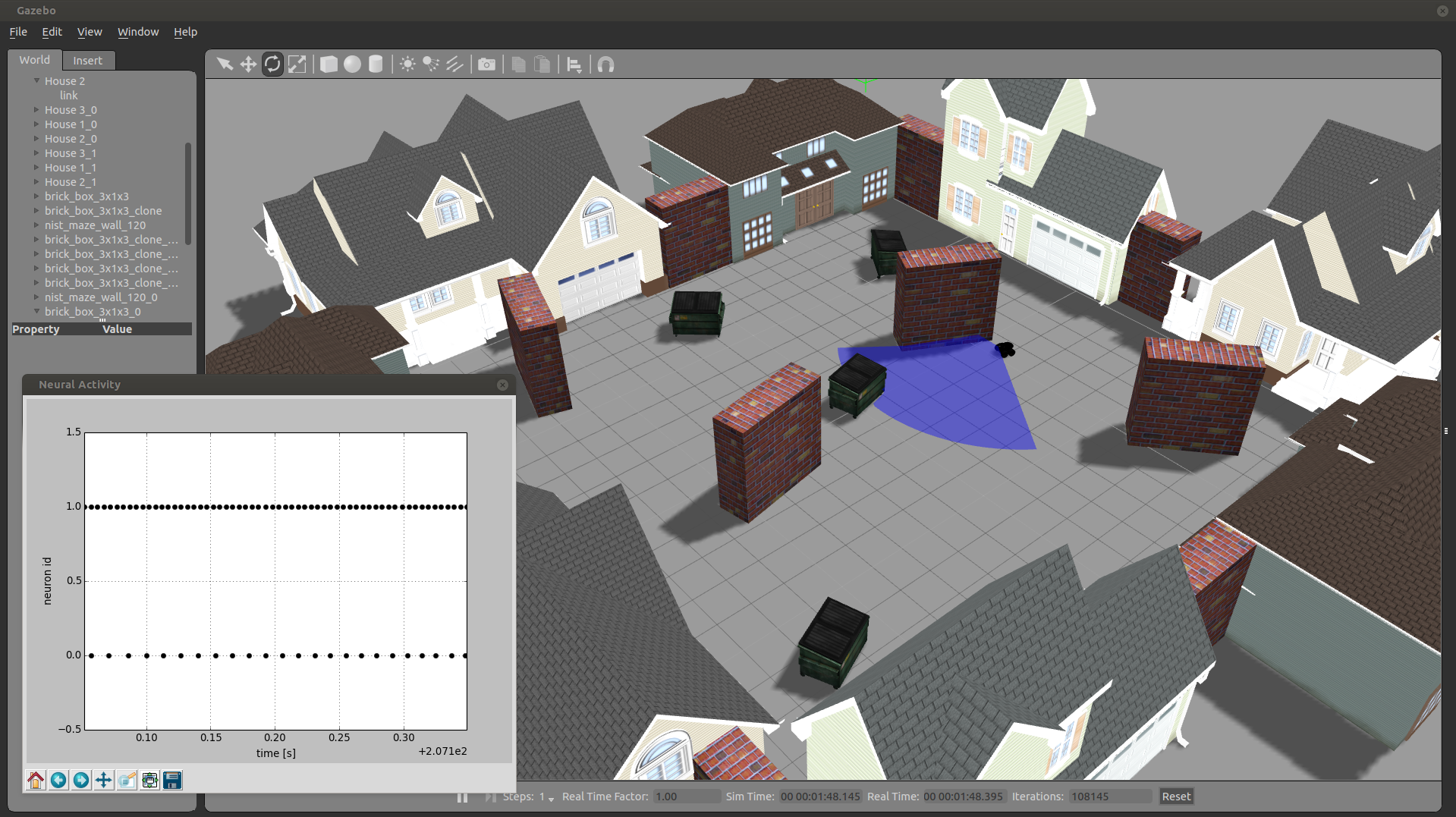}
\par\end{centering}

\caption{\label{fig:demo} A Braitenberg Vehicle simulated in Gazebo, controlled
by simulated neurons in NEST with the use of the ROS-MUSIC toolchain.
The inset shows the spiking activity of the two controlling neurons.}
\end{figure}
As an example for the usage of the toolchain, we created a Braitenberg
Vehicle III ``Explorer'' \citep{braitenberg1986vehicles} which
is simulated in the robotic environment Gazebo (see \prettyref{fig:demo}).
The Braitenberg Vehicle is implemented as a four-wheeled mobile robot
with an attached laser scanner for sensory input. With the use of
the ROS-MUSIC toolchain, two neurons, simulated in NEST, control this
vehicle to avoid obstacles. A video of the demonstration can be found
in the supplementary material and the source code is available on
GitHub%
\footnote{https://github.com/weidel-p/ros\_music\_adapter%
} .

\section{Discussion }

We have described our plug-in for MUSIC, which allows any neuronal
network simulator implementing a MUSIC interface to communicate with
any robotics simulators implementing a ROS interface. The plug-in
converts continuous signals from a robotics simulator into spike train
signals for a neuronal network simulator and vice versa.  We showed
that the toolchain allows real-time performance for a wide range of
configurations and provided a simple working example. In the following,
we discuss limits and perspectives for this approach.

\subsection*{Performance}

Dependent on the choice of encoder, our toolchain allows the simulation
of around $20,000$ neurons with a NEF encoder and up to $150,000$
neurons with a regular rate or Poisson encoder and achieve real-time
performance. Using the NEF encoder, it is recommended to have at least
$100$ neurons per input dimension in order to encode and decode stimuli
with root-mean-square errors of less than $1\%$ \citep{eliasmith2004neural},
meaning that our toolchain can encode a $200$-dimensional input with
NEF in real time on a single process of our hardware. 

If a higher dimensional input is required, there are two solutions
at hand. First, the NEF encoder could be parallelized and run in more
than one process. Second, if a less sophisticated but computationally
cheaper way of encoding is acceptable for the scientific question
at hand, we also provide regular rate and Poissonian encoding mechanisms.
A previous neurorobotic interface implemented the communication between
ROS and the neural simulators on the Python level \citep{hinkel_domain-specific_2015}.
Although no performance data were published in that study, it is clear
that the performance would be strongly limited by this approach. 

Another critical characteristic of the toolchain is the latency between
sensory input and motor command output of the toolchain, or in other
words the reaction time. To minimize the overhead and avoid repeated
communication of the same data, it makes sense to set the MUSIC time
step equal to the sensory update or motor command rate. However, \prettyref{fig:Bandwidth}
demonstrates that this leads to a rather high latency. The architecture
of communicating processes currently carries an unavoidable source
of latency, since data is buffered in MUSIC for every pair of communicating
processes. In any chain of communicating processes, the latency due
to buffering will thus grow in proportion to the length of the chain.
In future work, this can be tackled by combining multiple adapter,
encoder and decoder steps in a single process, thus simultaneously
minimizing communication and reaching a lower latency, and/or by introducing
non-buffered inter-process communication. Combining adapters, encoders
and decoders can be realised using a plug-in adapter API which maintains
independence between the software components.

\subsection*{Complexity of neuronal simulations}

In our measurements of the toolchain incorporating NEST and NEURON,
we used only minimal networks solving no particular task. However,
neural network models can rapidly become computationally complex,
especially when incorporating synaptic plasticity, large network sizes,
or multi-compartment neuron models. Such models would be impossible
to run in real-time using the current approach. In some cases this
issue could be solved by increasing the computational resources, using
a cluster or supercomputer \citep{helias2012supercomputers,kunkel2014spiking}.
However, this does not guarantee real-time execution, as neuronal
network simulators do not have perfect scaling, due to serial components
in their algorithms and communication overhead. Even if the problem
can theoretically be addressed in real-time by increasing resources,
it might not be feasible to access the quantity required. An alternative
prospect is offered by neuromorphic hardware, i.e. hardware purpose
built for simulation/emulation of neuronal networks. Two examples
that are being developed within the framework of the Human Brain Project
are SpiNNaker \citep{furber2013overview} and NM-PM1 \citep{schemmel2010wafer},
which have the potential to speed up the neuronal simulation massively.
In particular SpiNNaker has great potential for neurorobotic applications
and an interface between SpiNNaker and MUSIC is already in the prototype
phase. This development enhances the value of the toolchain we describe
for the neurorobotic community.

Whereas the real-time property is important for robotic control, it
is not nearly so important for addressing questions in the field of
computational neuroscience. Here, the advantage of the toolchain is
that a neuronal network simulation can be provided with rich sensory
input from an agent interacting with an environment that is easy for
the researcher to configure. In this case, arbitrarily computationally
demanding networks can be coupled with robotics simulators simply
by slowing down the latter to compensate - gazebo, for example, provides
a parameter to conveniently control the time scaling.

\subsection*{Applications}

The toolchain we describe gives researchers in computational neuroscience
the possibility to test their hypothesis and models under more realistic
conditions of noisy sensory input, and researchers in neurorobotics
the opportunity to investigate more realistic neurally based controllers.
One area of potential interest is the ability to construct interactions
between robotic simulators and hybrid neuronal simulations on multiple
scales, e.g. a network of point neuron models into which detailed
biophysical models are embedded, simulated by NEST and NEURON respectively.
This demonstrates a further advantage of the MUSIC-based interaction
over pairings of particular simulators or the Python-based interaction
presented by \citet{hinkel_domain-specific_2015}. 

Moreover, our toolchain is particularly well suited for studying closed-loop
scenarios, where the neural network receives stimuli from a complex
environment and produces an output, which in turn causes the robotic
agent to perform actions within that environment. For example, a robotic
agent can be placed in a classic experimental set-up like a T-maze
and the behaviour of the robot adapted by a neurally implemented reinforcement
learner \citep{jitsev2012learning,potjans2011imperfect,fremaux2013reinforcement,friedrich2015goal}.
Here, there is a clear advantage over studying such questions just
using neural simulators, as the representation of an external environment
as a collection of neural recorders and stimulators is complex, and
difficult to either generalize or customize. By separating the concerns
of environmental, motor, and sensory representation from those of
neural processing, our toolchain provides a highly flexible and performant
research approach.

\section*{Acknowledgments}

We thank Karolína Korvasová, Tom Tetzlaff and Jenia Jitsev for helpful
discussions, and Jenia also for his comments on an earlier version
of this manuscript. We acknowledge partial support by the German Federal
Ministry of Education through our German-Japanese Computational Neuroscience
Project (BMBF Grant 01GQ1343), EuroSPIN, the Helmholtz Alliance through
the Initiative and Networking Fund of the Helmholtz Association and
the Helmholtz Portfolio theme ``Supercomputing and Modeling for the
Human Brain'' and the European Union Seventh Framework Programme
(FP7/2007-2013) under grant agreement no. 604102 (HBP). We also thank
the INCF for travel support enabling work meetings.

\bibliographystyle{neuralcomput_natbib}
\bibliography{papers}

\begin{thebibliography}{}

\bibitem[\protect\citeauthoryear{Antonelo et~al.}{Antonelo
  et~al.}{2007}]{antonelo_event_2007}
Antonelo, E.~A., Schrauwen, B., Dutoit, X., Stroobandt, D., \& Nuttin, M.
  (2007).
\newblock Event {Detection} and {Localization} in {Mobile} {Robot} {Navigation}
  {Using} {Reservoir} {Computing}.
\newblock In J.~M. de~S\'a, L.~A. Alexandre, W.~Duch, \& D.~Mandic (Eds.), {\em
  Artificial {Neural} {Networks} -- {ICANN} 2007}, Volume 4669 of {\em Lecture
  {Notes} in {Computer} {Science}}, Berlin, Heidelberg, pp.\  660--669.
  Springer Berlin Heidelberg.

\bibitem[\protect\citeauthoryear{Averbeck}{Averbeck}{2009}]{Averbeck2009}
Averbeck, B.~B. (2009).
\newblock {Poisson or not Poisson: differences in spike train statistics
  between parietal cortical areas.}
\newblock {\em Neuron\/}~{\em 62\/}(3), 310--1.

\bibitem[\protect\citeauthoryear{Braitenberg}{Braitenberg}{1986}]{braitenberg1986vehicles}
Braitenberg, V. (1986).
\newblock {\em Vehicles: Experiments in synthetic psychology}.
\newblock MIT press.

\bibitem[\protect\citeauthoryear{Calvin \& Stevens}{Calvin \&
  Stevens}{1967}]{Calvin1967}
Calvin, W.~H., \& Stevens, C.~F. (1967).
\newblock {Synaptic noise as a source of variability in the interval between
  action potentials.}
\newblock {\em Science (New York, N.Y.)\/}~{\em 155\/}(3764), 842--4.

\bibitem[\protect\citeauthoryear{Chorley \& Seth}{Chorley \&
  Seth}{2008}]{chorley2008closing}
Chorley, P., \& Seth, A.~K. (2008).
\newblock Closing the sensory-motor loop on dopamine signalled reinforcement
  learning.
\newblock In {\em From Animals to Animats 10}, pp.\  280--290. Springer.

\bibitem[\protect\citeauthoryear{Dasgupta, W\"org\"otter, \&
  Manoonpong}{Dasgupta et~al.}{2013}]{dasgupta_information_2013}
Dasgupta, S., W\"org\"otter, F., \& Manoonpong, P. (2013).
\newblock Information dynamics based self-adaptive reservoir for delay temporal
  memory tasks.
\newblock {\em Evolving Systems\/}~{\em 4\/}(4), 235--249.

\bibitem[\protect\citeauthoryear{Dayan \& Abbott}{Dayan \&
  Abbott}{2001}]{Dayan2001}
Dayan, P., \& Abbott, L.~F. (2001).
\newblock {\em {Theoretical Neuroscience: Computational and Mathematical
  Modeling of Neural Systems}}.

\bibitem[\protect\citeauthoryear{Djurfeldt et~al.}{Djurfeldt
  et~al.}{2010}]{djurfeldt2010run}
Djurfeldt, M., Hjorth, J., Eppler, J.~M., Dudani, N., Helias, M., Potjans,
  T.~C., Bhalla, U.~S., Diesmann, M., Kotaleski, J.~H., \& Ekeberg, {\"O}.
  (2010).
\newblock Run-time interoperability between neuronal network simulators based
  on the music framework.
\newblock {\em Neuroinformatics\/}~{\em 8\/}(1), 43--60.

\bibitem[\protect\citeauthoryear{Eliasmith \& Anderson}{Eliasmith \&
  Anderson}{2004}]{eliasmith2004neural}
Eliasmith, C., \& Anderson, C.~H. (2004).
\newblock {\em Neural engineering: Computation, representation, and dynamics in
  neurobiological systems}.
\newblock MIT press.

\bibitem[\protect\citeauthoryear{Eppler et~al.}{Eppler
  et~al.}{2015}]{eppler_2015_32969}
Eppler, J.~M., Pauli, R., Peyser, A., Ippen, T., Morrison, A., Senk, J.,
  Schenck, W., Bos, H., Helias, M., Schmidt, M., Kunkel, S., Jordan, J.,
  Gewaltig, M.-O., Bachmann, C., Schuecker, J., Albada, S., Zito, T., Deger,
  M., Michler, F., Hagen, E., Setareh, H., Riquelme, L., Shirvani, A., Duarte,
  R., Deepu, R., \& Plesser, H.~E. (2015).
\newblock Nest 2.8.0.

\bibitem[\protect\citeauthoryear{Fr\'emaux, Sprekeler, \& Gerstner}{Fr\'emaux
  et~al.}{2013}]{fremaux_reinforcement_2013}
Fr\'emaux, N., Sprekeler, H., \& Gerstner, W. (2013).
\newblock Reinforcement {Learning} {Using} a {Continuous} {Time}
  {Actor}-{Critic} {Framework} with {Spiking} {Neurons}.
\newblock {\em PLoS Computational Biology\/}~{\em 9\/}(4), e1003024.

\bibitem[\protect\citeauthoryear{Fr{\'e}maux, Sprekeler, \&
  Gerstner}{Fr{\'e}maux et~al.}{2013}]{fremaux2013reinforcement}
Fr{\'e}maux, N., Sprekeler, H., \& Gerstner, W. (2013).
\newblock Reinforcement learning using a continuous time actor-critic framework
  with spiking neurons.
\newblock {\em PLoS Comput Biol\/}~{\em 9\/}(4), e1003024.

\bibitem[\protect\citeauthoryear{Friedrich \& Lengyel}{Friedrich \&
  Lengyel}{2015}]{friedrich2015goal}
Friedrich, J., \& Lengyel, M. (2015).
\newblock Goal-directed decision making with spiking neurons.

\bibitem[\protect\citeauthoryear{Furber et~al.}{Furber
  et~al.}{2013}]{furber2013overview}
Furber, S.~B., Lester, D.~R., Plana, L.~A., Garside, J.~D., Painkras, E.,
  Temple, S., \& Brown, A.~D. (2013).
\newblock Overview of the spinnaker system architecture.
\newblock {\em Computers, IEEE Transactions on\/}~{\em 62\/}(12), 2454--2467.

\bibitem[\protect\citeauthoryear{Gewaltig \& Diesmann}{Gewaltig \&
  Diesmann}{2007}]{Gewaltig:NEST}
Gewaltig, M.-O., \& Diesmann, M. (2007).
\newblock Nest (neural simulation tool).
\newblock {\em Scholarpedia\/}~{\em 2\/}(4), 1430.

\bibitem[\protect\citeauthoryear{Hanuschkin et~al.}{Hanuschkin
  et~al.}{2010}]{Hanuschkin2010}
Hanuschkin, A., Kunkel, S., Helias, M., Morrison, A., \& Diesmann, M. (2010).
\newblock {A general and efficient method for incorporating precise spike times
  in globally time-driven simulations.}
\newblock {\em Frontiers in neuroinformatics\/}~{\em 4\/}(October), 113.

\bibitem[\protect\citeauthoryear{Helias et~al.}{Helias
  et~al.}{2012}]{helias2012supercomputers}
Helias, M., Kunkel, S., Masumoto, G., Igarashi, J., Eppler, J.~M., Ishii, S.,
  Fukai, T., Morrison, A., \& Diesmann, M. (2012).
\newblock Supercomputers ready for use as discovery machines for neuroscience.
\newblock {\em Front Neuroinform\/}~{\em 6\/}(26), 2.

\bibitem[\protect\citeauthoryear{Hines \& Carnevale}{Hines \&
  Carnevale}{2001}]{hines2001neuron}
Hines, M., \& Carnevale, N.~T. (2001).
\newblock Neuron: a tool for neuroscientists.
\newblock {\em The Neuroscientist\/}~{\em 7\/}(2), 123--135.

\bibitem[\protect\citeauthoryear{Hinkel et~al.}{Hinkel
  et~al.}{2015}]{hinkel_domain-specific_2015}
Hinkel, G., Groenda, H., Vannucci, L., Denninger, O., Cauli, N., \& Ulbrich, S.
  (2015).
\newblock A {Domain}-{Specific} {Language} ({DSL}) for {Integrating} {Neuronal}
  {Networks} in {Robot} {Control}.
\newblock In {\em Proceedings of the 2015 {Joint} {MORSE}/{VAO} {Workshop} on
  {Model}-{Driven} {Robot} {Software} {Engineering} and {View}-based
  {Software}-{Engineering}}, {MORSE}/{VAO} '15, New York, NY, USA, pp.\  9--15.
  ACM.

\bibitem[\protect\citeauthoryear{Jitsev, Morrison, \& Tittgemeyer}{Jitsev
  et~al.}{2012}]{jitsev2012learning}
Jitsev, J., Morrison, A., \& Tittgemeyer, M. (2012).
\newblock Learning from positive and negative rewards in a spiking neural
  network model of basal ganglia.
\newblock In {\em Neural Networks (IJCNN), The 2012 International Joint
  Conference on}, pp.\  1--8. IEEE.

\bibitem[\protect\citeauthoryear{Kunkel et~al.}{Kunkel
  et~al.}{2014}]{kunkel2014spiking}
Kunkel, S., Schmidt, M., Eppler, J.~M., Plesser, H.~E., Masumoto, G., Igarashi,
  J., Ishii, S., Fukai, T., Morrison, A., Diesmann, M., et~al. (2014).
\newblock Spiking network simulation code for petascale computers.
\newblock {\em Frontiers in neuroinformatics\/}~{\em 8\/}(78).

\bibitem[\protect\citeauthoryear{Legenstein \& Maass}{Legenstein \&
  Maass}{2014}]{legenstein2014ensembles}
Legenstein, R., \& Maass, W. (2014).
\newblock Ensembles of spiking neurons with noise support optimal probabilistic
  inference in a dynamically changing environment.
\newblock {\em PLoS Comput Biol\/}~{\em 10\/}(10), e1003859.

\bibitem[\protect\citeauthoryear{Moren, Sugimoto, \& Doya}{Moren
  et~al.}{2015}]{moren2015real}
Moren, J., Sugimoto, N., \& Doya, K. (2015).
\newblock Real-time utilization of system-scale neuroscience models.
\newblock {\em Journal of the Japanese Neural Network Society\/}~{\em 22\/}(3),
  125--132.

\bibitem[\protect\citeauthoryear{Morrison, Aertsen, \& Diesmann}{Morrison
  et~al.}{2007}]{Morrison2007}
Morrison, A., Aertsen, A., \& Diesmann, M. (2007).
\newblock {Spike-timing-dependent plasticity in balanced random networks.}
\newblock {\em Neural computation\/}~{\em 19\/}(6), 1437--67.

\bibitem[\protect\citeauthoryear{Potjans, Diesmann, \& Morrison}{Potjans
  et~al.}{2011}]{potjans2011imperfect}
Potjans, W., Diesmann, M., \& Morrison, A. (2011).
\newblock An imperfect dopaminergic error signal can drive temporal-difference
  learning.
\newblock {\em PLoS Comput Biol\/}~{\em 7\/}(5), e1001133.

\bibitem[\protect\citeauthoryear{Qui\~nonez et~al.}{Qui\~nonez
  et~al.}{2015}]{quinonez_autonomous_2015}
Qui\~nonez, Y., Ramirez, M., Lizarraga, C., Tostado, I., \& Bekios, J. (2015).
\newblock Autonomous {Robot} {Navigation} {Based} on {Pattern} {Recognition}
  {Techniques} and {Artificial} {Neural} {Networks}.
\newblock In J.~M.~F. Vicente, J.~R. \'Alvarez-S\'anchez, F.~d. l.~P. L\'opez,
  F.~J. Toledo-Moreo, \& H.~Adeli (Eds.), {\em Bioinspired {Computation} in
  {Artificial} {Systems}}, Number 9108 in Lecture {Notes} in {Computer}
  {Science}, pp.\  320--329. Springer International Publishing.

\bibitem[\protect\citeauthoryear{Quigley et~al.}{Quigley
  et~al.}{2009}]{quigley2009ros}
Quigley, M., Conley, K., Gerkey, B., Faust, J., Foote, T., Leibs, J., Wheeler,
  R., \& Ng, A.~Y. (2009).
\newblock Ros: an open-source robot operating system.
\newblock In {\em ICRA workshop on open source software}, Volume~3, pp.\ ~5.

\bibitem[\protect\citeauthoryear{Rotter \& Diesmann}{Rotter \&
  Diesmann}{1999}]{Rotter1999}
Rotter, S., \& Diesmann, M. (1999).
\newblock {Exact digital simulation of time-invariant linear systems with
  applications to neuronal modeling.}
\newblock {\em Biological cybernetics\/}~{\em 81\/}(5-6), 381--402.

\bibitem[\protect\citeauthoryear{Schemmel et~al.}{Schemmel
  et~al.}{2010}]{schemmel2010wafer}
Schemmel, J., Bruderle, D., Grubl, A., Hock, M., Meier, K., \& Millner, S.
  (2010).
\newblock A wafer-scale neuromorphic hardware system for large-scale neural
  modeling.
\newblock In {\em Circuits and systems (ISCAS), proceedings of 2010 IEEE
  international symposium on}, pp.\  1947--1950. IEEE.

\end{thebibliography}

\end{document}